\documentclass[runningheads]{llncs}
\usepackage[T1]{fontenc}
\usepackage{graphicx}
\usepackage{xcolor}
\usepackage{xspace}
\usepackage{amsmath}
\usepackage{bbm} % For one vector
\usepackage[labelformat=simple]{subcaption}
\usepackage{algorithm}
\usepackage{algpseudocode}
\usepackage{mdwlist}
\usepackage{multirow} % For table
\usepackage{cuted}
\usepackage{afterpage}
\usepackage{stfloats}
\usepackage{hhline}
\usepackage{tabularx}
\usepackage{amsthm}
\usepackage{enumitem}
\usepackage[skip=4pt]{caption}

\usepackage{pifont}
\usepackage{balance}

\usepackage{amsfonts}
\usepackage{hyperref}
\usepackage{booktabs}
\usepackage{mathtools}

\newcommand{\set}[1]{#1}
\newcommand{\mat}[1]{\mathbf{#1}}
\newcommand{\vect}[1]{\mathbf{#1}}
\newcommand{\Graph}{\mathcal{G}}
\newcommand{\graph}[1]{\Graph_{#1}}

\newcommand{\gvect}[1]{\boldsymbol{#1}}
\newcommand{\gmat}[1]{\boldsymbol{#1}}

\newcommand{\OriGraph}{\Graph}
\newcommand{\AugGraph}{\Graph^{+}}

\newcommand{\OriZ}{\vect{z}_{\OriGraph}}
\newcommand{\AugZ}{\vect{z}_{\AugGraph}}

\newcommand{\nodes}{\mathcal{V}}
\newcommand{\edges}{\mathcal{E}}

\newcommand{\augmenter}{\mathcal{T}_{p}}
\newcommand{\encoder}{f_{\theta}}
\newcommand{\classifier}{g_{\phi}}
\newcommand{\neunet}{h_{\omega}}

\newcommand{\Laware}{\mathcal{L}_{\texttt{aware}}}
\newcommand{\Lcr}{\mathcal{L}_{\texttt{cr}}}
\newcommand{\Lori}{\mathcal{L}_{\texttt{base}}}
\newcommand{\Laugward}{\mathcal{L}_{\texttt{AugWard}}}

\newcommand{\LambdaAware}{\lambda_{\texttt{aware}}}
\newcommand{\LambdaCR}{\lambda_{\texttt{cr}}}

\newcommand{\OriPred}{\vect{p}_{\OriGraph}}
\newcommand{\AugPred}{\vect{p}_{\AugGraph}}

\definecolor{ForestGreen}{rgb}{0.133, 0.545, 0.133}
\definecolor{torange}{rgb}{0.949, 0.522, 0.}
\newcommand{\method}{\textsc{AugWard}\xspace}
\newcommand{\methodfullbold}{\textbf{\underline{Aug}}mentation-A\textbf{\underline{war}}e Training with Graph \textbf{\underline{D}}istance and Consistency Regularization\xspace}
\newcommand{\githublink}{\url{https://github.com/snudatalab/AugWard}\xspace}

\newcolumntype{Y}{>{\centering\arraybackslash}X}
\newtheoremstyle{dotless}{}{}{\itshape}{}{\bfseries}{}{ }{}

\theoremstyle{dotless}

\newcommand{\dd}{\texttt{DD}\xspace}
\newcommand{\enzymes}{\texttt{ENZYMES}\xspace}
\newcommand{\imdbb}{\texttt{IMDB-BINARY}\xspace}
\newcommand{\imdbm}{\texttt{IMDB-MULTI}\xspace}
\newcommand{\nciOne}{\texttt{NCI1}\xspace}
\newcommand{\nciNine}{\texttt{NCI109}\xspace}
\newcommand{\proteins}{\texttt{PROTEINS}\xspace}
\newcommand{\ptcmr}{\texttt{PTC-MR}\xspace}
\newcommand{\redditb}{\texttt{REDDIT-BINARY}\xspace}
\newcommand{\twitter}{\texttt{TWITTER}\xspace}
\newcommand{\zinc}{\texttt{ZINC15}\xspace}

\newcommand{\enzymesShort}{\texttt{ENZ}\xspace}
\newcommand{\imdbbShort}{\texttt{I-B}\xspace}
\newcommand{\imdbmShort}{\texttt{I-M}\xspace}
\newcommand{\proteinsShort}{\texttt{PRO}\xspace}
\newcommand{\ptcmrShort}{\texttt{PTC}\xspace}
\newcommand{\redditbShort}{\texttt{R-B}\xspace}
\newcommand{\twitterShort}{\texttt{TWI}\xspace}

\let\llncssubparagraph\subparagraph
\let\subparagraph\paragraph
\usepackage{titlesec}
\let\subparagraph\llncssubparagraph
\titlespacing{\section}{0pt}{*1}{*1}
\titlespacing{\subsection}{0pt}{*1}{*0.5}

\newcommand{\smallsection}[1]{\vspace{1mm}\noindent\smash{\textbf{#1.}}}
\setlist[itemize]{topsep=1pt}

\AtBeginDocument{\setlength\abovedisplayskip{4pt}}
\AtBeginDocument{\setlength\belowdisplayskip{4pt}}

\AtBeginDocument{\setlength{\textfloatsep}{2pt}}
\AtBeginDocument{\setlength{\abovecaptionskip}{2pt}}
\AtBeginDocument{\setlength{\belowcaptionskip}{2pt}}

\graphicspath{ {images/} }

\begin{document}
\title{\method: Augmentation-Aware Representation Learning for Accurate Graph Classification}
\titlerunning{\method}
\author{
 Minjun Kim\inst{1}
% \orcidID{0009-0007-1119-5615}
 \and
 Jaehyeon Choi\inst{1}
% \orcidID{0000-0002-7214-5554}
 \and
 SeungJoo Lee\inst{1}
% \orcidID{0000-0003-2657-411X}
 \and
 Jinhong Jung\inst{2}\thanks{Corresponding Authors.}
% \orcidID{0000-0002-5533-1507}
 \and
 U Kang\inst{1}$^*$
% \orcidID{0000-0002-8774-6950}
%Anonymous Author(s)
}
\authorrunning{
 M. Kim et al.
%Anonymous Author(s)
}
\institute{
 Seoul National University, Seoul, South Korea \\
 \email{\{minjun.kim,jaehyeon\_choi,hera0131,ukang\}@snu.ac.kr} \\ \and
 Soongsil University, Seoul, South Korea \\
 \email{jinhong@ssu.ac.kr}
%Anonymous Institute(s)\\
%\email{anonymous.email(s)@domain}
}
\maketitle              % typeset the header of the contribution

\begin{abstract}
% 169/200 words
% https://wordcounter.net/
How can we accurately classify graphs?
Graph classification is a pivotal task in data mining with applications in social network analysis, web analysis, drug discovery, molecular property prediction, etc.
Graph neural networks have achieved the state-of-the-art performance in graph classification, but they consistently struggle with overfitting.
To mitigate overfitting, researchers introduced various representation learning methods utilizing graph augmentation.
However, existing methods rely on simplistic use of graph augmentation, which loses augmentation-induced differences and limits the expressiveness of representations.

In this paper, we propose \textbf{\method} (\methodfullbold), a novel graph representation learning framework that carefully considers the diversity introduced by graph augmentation.
\method applies augmentation-aware training to predict the graph distance between the augmented graph and its original one, aligning the representation difference directly with graph distance at both feature and structure levels.
Furthermore, \method employs consistency regularization to encourage the classifier to handle richer representations.
Experimental results show that \method gives the state-of-the-art performance in supervised, semi-supervised graph classification, and transfer learning. 
\end{abstract}
\section{Introduction}
%\postsecMargin
\label{sec:intro}
% ================ Figure 1: Augmentation-aware Training ==================
\begin{figure}[t]
	\centering
	\includegraphics[width=0.95\linewidth]{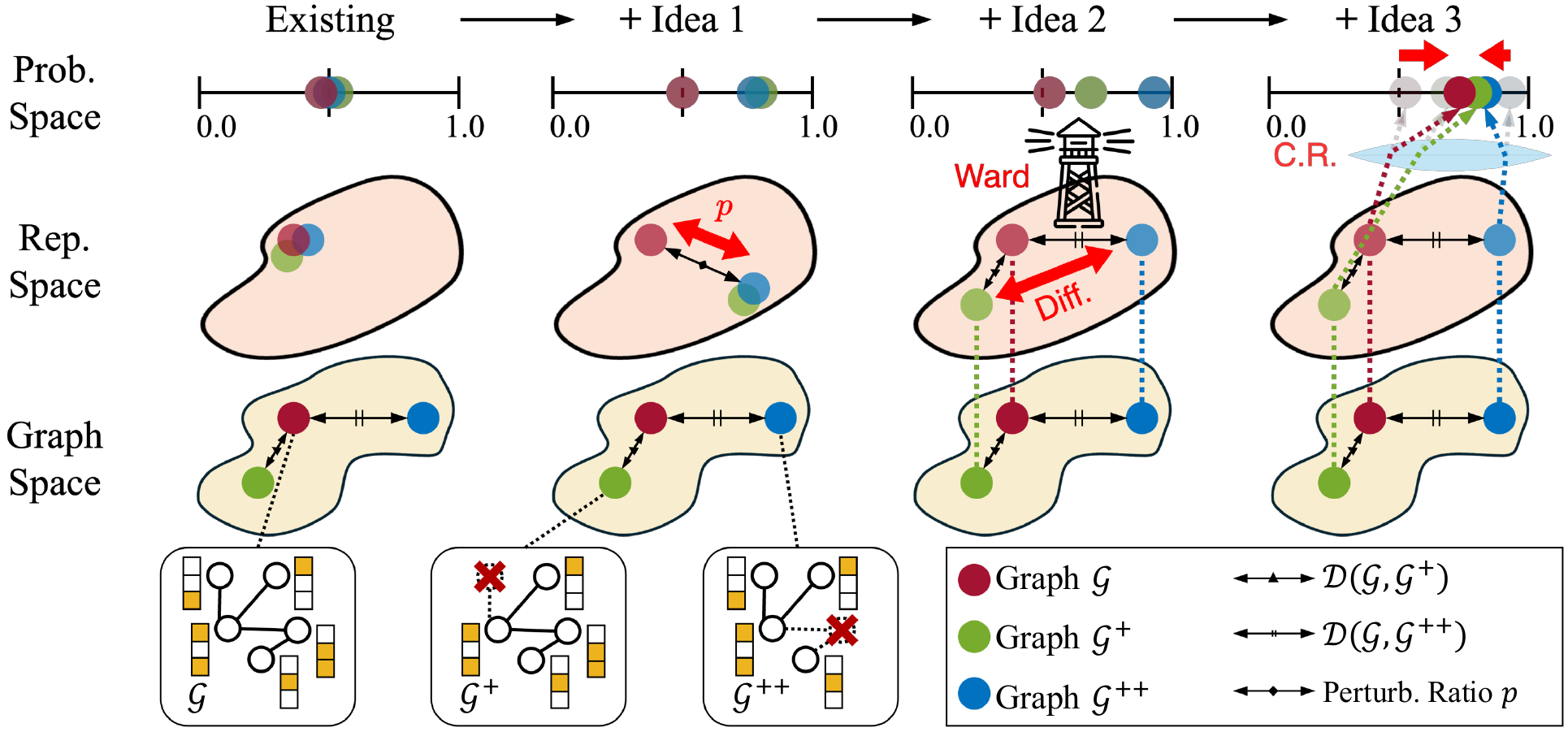}
	\vspace{2mm}
	\caption{
        Main ideas of \method\ are:
        I1) augmentation-aware training,
        I2) graph distance-based difference, and
        I3) Consistency Regularization (C.R.).
        % Refer to Section~\ref{sec:intro} for details.
    }
	\label{fig:aware}
	%\vspace{-2mm}
\end{figure}
% ================ END ===========================================

%%%% Paragraph 1. Problem + Importance
% \textit{How can we accurately classify graphs?}
%
Graph classification~\cite{TAG} is an important data mining task that aims to classify a graph into predefined classes based on its structural properties and features.
This task has been attracting attention as it enables reliable predictions from graph-structured data in various applications.
% ~\cite{} % reduce the number of citations
% such as
% social network analysis~\cite{FairComp},
% web mining~\cite{GraphArticle},
% recommender systems~\cite{GraphClassRec},
% bioinformatics~\cite{GraphOnBiology},
% cheminformatics~\cite{GraphOnChemistry}, etc.
% computer vision~\cite{LSP}.
%
%%%% Paragraph 2. Previous Approaches
Previous studies~\cite{GraphAug,GraphTransplant} have consistently shown the advantages of Graph Neural Networks (GNNs)~\cite{SignedJung,GCN} over traditional methods based on graph kernels.
GNNs capture higher-order structures through end-to-end multi-layered architectures, overcoming the limited expressiveness due to hand-crafted features of graph kernels.
With the rise of GNNs, graph augmentation~\cite{S-Mixup,NodeSam} has become crucial to overcome the overfitting issue.
Recent approaches further enhance training for better generalization under unsupervised or semi-supervised settings, such as contrastive learning~\cite{CuCo,GraphCL} or mutual information maximization~\cite{InfoGraph}.
Given this background, the efficient use of graph augmentation is a key aspect for graph classification. 

%%%% Paragraph 3. Difficulties and Challenges
%
However, the state-of-the-art methods~\cite{CuCo,GCL-SPAN,GraphCL} are suboptimal due to two major limitations by simplistic adaptation of graph augmentation.
First, the difference between augmented graphs and their originals in terms of structures and features, termed the \textit{augmentation-induced difference}, is largely ignored in existing methods (see Figure~\ref{fig:nonlinear}).
They aim to predict consistent labels for augmented graphs and their originals without capturing variations, leading to augmentation-invariant representations.
Second, existing methods rely on the deceptive assumption that the perturbation ratio $p$ ensures similarity among augmented graphs.
Graph augmentation techniques~\cite{S-Mixup,NodeSam} randomly modify graphs according to $p$ (or the extent of change), assuming that augmented graphs with the same $p$ are similar.
However, this misleading assumption overlooks the significant diversity among the augmented graphs generated at a fixed $p$ (see Figure~\ref{fig:randomness}), limiting previous designs from adequately considering the difference.
%
%\begin{itemize}[leftmargin=4.5mm]
%    \item{
%        \textbf{Ignorance of the difference between original and augmented graphs.}
%        Existing methods disregard this difference in either structure or feature, called \textit{augmentation-induced difference} (see Figure~\ref{fig:nonlinear}).
%        They aim to predict consistent labels for augmented graphs and their originals without capturing variations, leading to augmentation-invariant representations.
%        % introduced by augmentation in graph representations.
%    }
%    \item{
%        \textbf{Deceptiveness of perturbation ratio $p$.}
%        Existing graph augmentation techniques~\cite{S-Mixup,NodeSam} randomly modify original graphs according to $p$ (or the extent of change), assuming that augmented graphs with same $p$ are similar.
%        However, this misleading assumption overlooks the significant diversity among the augmented graphs generated at a fixed $p$ (see Figure~\ref{fig:randomness}), limiting previous designs from adequately considering the difference.
%        % the inherent randomness of graph augmentation
%    }
%\end{itemize}

%%%% Paragraph 4. Main contribution + ideas
We propose \textbf{\method}, a novel framework for graph representation learning designed to effectively address these limitations.
Figure~\ref{fig:aware} depicts our ideas.
We perform augmentation-aware training to enrich graph representations, by aligning the difference between augmented graphs and their original with that estimated by their representations (Idea 1).
Observing that the perturbation ratio $p$ is inappropriate to represent graph-level differences, we explicitly measure the graph distance between original and augmented graphs, considering both structure and features with Fused Gromov-Wasserstein distance (Idea 2).
Furthermore, we exploit consistency regularization to ensure similar predictions from distinguishable representations of original and augmented graphs (Idea 3).
%The ideas of \method are encapsulated in the two loss functions for each augmentation-aware training with graph distance and consistency regularization.

\method is powerful and versatile since it is easily integrated with any method utilizing graph augmentation, enhancing their accuracy across various learning settings, such as supervised, semi-supervised, and transfer learning.
%
%%%% Paragraph 5. List of contributions (w/ detail)
Our main contributions are summarized as follows:
\begin{itemize}[leftmargin=4.5mm]
    \item{
        \textbf{Observation.}
        We empirically observe that existing methods disregard the augmentation-induced difference and an appropriate metric is required to accurately measure this difference (See Figures~\ref{fig:nonlinear} and~\ref{fig:randomness}).
	}
    \item{
        \textbf{Framework.}
        We propose \method, a novel graph representation learning framework that enriches the graph representations by the efficient use of graph augmentation.
        \method incorporates augmentation-aware training with graph distance and consistency regularization (See Figures~\ref{fig:aware} and~\ref{fig:method}).
    }
    \item{
        \textbf{Experiments.}
        \method consistently elevates the classification accuracy of the state-of-the-art methods in supervised, semi-supervised graph classification, and transfer learning tasks (see Tables~\ref{tab:q1},~\ref{tab:q2}, and~\ref{tab:q3}).
        % The success of \method in transfer learning applications further validates the high quality of its representations.
    }
\end{itemize}
%
%\smallsection{Reproducibility}
The source code of \method and the datasets are available at
\githublink.
% 
%\presecMargin

\section{Preliminaries}
%\postsecMargin
\label{sec:prelim}
% We describe the graph classification problem and the preliminaries.

% \presubsecMargin
%%% Subsection 2.1. Problem Definition
% \subsection{Problem Definition}
\smallsection{Problem Definition}
% \label{subsec:2.1}
% \postsubsecMargin
%
Graph classification predicts the label of a graph $\Graph = (\nodes, \edges, \mat{X})$, where $\nodes$ and $\edges$ are the sets of nodes and edges, respectively.
$\mat{X} \in \mathbb{R}^{n \times d}$ is the feature matrix where $n = |\nodes|$ and $d$ is the number of features.
\vspace{-1mm}
\begin{problem}[Graph Classification]
{\color{white}whitespace}
\label{prob:mbr}
\begin{itemize}[leftmargin=4mm]
    \item {
        \textbf{Input}: a set $\set{G} = \{\graph{1}, \cdots, \graph{N}\}$ of $N$ distinct graphs and a set $\set{Y} = \{y_{1}, \cdots, y_{N}\}$ of labels where each label $y_{i}$ corresponds to $\graph{i}$ and belongs to a set $\set{C}$ of classes.
    }
    \item {
        \textbf{Output}: prediction probabilities $P(y|\graph{i})$ that each graph $\graph{i}$ is associated with the label $y \in \set{C}$.
        %\hspace{-1.5mm}
%        \footnote{
%            We assume that class labels are represented as integers between $1$ and  $|\set{C}|$, i.e., $P(y = i | \graph{G})$ indicates the probability that label $y$ of $\graph{G}$ belongs to class $i$.
%        }
    }
\end{itemize}
\end{problem}
%
% \presubsecMargin
% %%%% Subsection 2.2. Graph Neural Networks
% \subsection{Graph Neural Networks}
% \label{subsec:2.2}
% \postsubsecMargin

\smallsection{Graph Neural Networks}
Graph Neural Networks (GNNs)~\cite{GIN,BPN} are deep learning models designed to process graph-structured data by leveraging message passing algorithm between nodes to learn their representations.
They jointly train an encoder $\encoder$ and a classifier $\classifier$ for the graph classification task.
The encoder $\encoder$ transforms a given graph $\OriGraph$ into its representation $\OriZ \in \mathbb{R}^{d}$ using learnable parameters $\theta$, where $d$ is the embedding dimension.
$\encoder$ consists of $L$ layers, where the message-passing step of the $l$-th layer is as follows:
\begin{equation}
    \label{eq:eq1}
        \vect{\tilde{h}}^{(l)}_{u} \leftarrow \texttt{AGGREGATE}\Big( \big\{\textbf{h}^{(l-1)}_{v} \colon v \in \mathcal{N}_{u} \big\} \Big),~
        \vect{h}^{(l)}_{u} \leftarrow \texttt{COMBINE}\Big(\vect{h}^{(l-1)}_{u},  \vect{\tilde{h}}^{(l)}_{u}\Big),
    \vspace{-1mm}
\end{equation}
where $\vect{h}^{(l)}_{u}$ is the hidden embedding of node $u$ at layer $l$, and $\mathcal{N}_{u}$ is the set of neighbors for node $u$.
$\vect{h}^{(0)}_{u}$ is initialized with $\vect{x}_{u}$, the $u$-th row vector of $\mat{X}$.
After $L$ propagation, $\OriZ$ is obtained as follows, where $[L] \coloneq \{0, \cdots, L\}$:
\begin{equation}
    \label{eq:eq2}
        \OriZ \leftarrow \texttt{READOUT}\Big(\big\{\vect{h}^{(l)}_{u} \: | \: u \in \nodes, l \in [L] \big\}\Big).
\end{equation}
Given $\OriZ$, the classifier $\classifier$ produces a vector $\OriPred \in \mathbb{R}^{|C|}$ of predicted probabilities, i.e., $\OriPred \leftarrow \classifier(\OriZ)$, where the $i$-th entry of $\OriPred$ is $P(y = i | \OriGraph)$. In this paper, we assume that class labels are represented as integers between $1$ and  $|\set{C}|$, i.e., $P(y = i | \graph{G})$ indicates the probability that label $y$ of $\graph{G}$ belongs to class $i$.
%
% Different choices for the above functions lead to diverse GNN designs~\cite{GraphSAGE,GCN}.
% Even though all GNNs fit within our framework, we exploit GIN~\cite{GIN} as the default GNN due to its state-of-the-art performance.
% For the \texttt{READOUT} function, we sum the hidden embeddings of every node across all layers as in~\cite{GraphMAE,DualGraph}.
% Our method is also effective for other GNNs, with empirical results provided in the Supplementary Materials~\cite{supp}.

% \presubsecMargin
%%%% Subsection 2.3. Graph Augmentation
% \subsection{Graph Augmentation}
% \label{subsec:2.3}
\smallsection{Graph Augmentation}
% \postsubsecMargin
%
Graph augmentation aims to increase the volume of graph instances by modifying the original graphs while retaining their labels.
Let $\augmenter(\cdot | \OriGraph)$ represent the augmentation distribution conditioned on the original graph $\OriGraph$, where the hyperparameter $p$ denotes the perturbation ratio indicating the amount of change from $\OriGraph$.
Then, an augmented graph $\AugGraph$ is randomly sampled as $\AugGraph \sim \augmenter\big(\AugGraph \: | \: \OriGraph\big).$
Various designs have been proposed for $\augmenter$, such as drop-based methods~\cite{NodeAug,GraphCL} or mixup-based methods~\cite{FGWMixup,GraphTransplant}.
The former randomly either remove or mask attributes of nodes and edges according to the ratio $p$.
The latter fuse two source graphs by the ratio $p$, with the resulting graph having a linearly interpolated label from the original graphs.

% \presubsecMargin
%%%% Subsection 2.4. Fused Gromov-Wasserstein Distance
% \subsection{Fused Gromov-Wasserstein Distance}
% \label{subsec:2.4}
\smallsection{Fused Gromov-Wasserstein Distance}
% \postsubsecMargin
%
Fused Gromov-Wasserstein Distance (FGWD)~\cite{FGWMixup,FGW} measures the distance between two graphs by combining both feature-level and structure-level differences, inspired from the optimal transport problem.
FGWD employs Wasserstein distance $\texttt{WD}(\cdot, \cdot, \cdot)$ to capture feature-level differences and Gromov-Wasserstein distance $\texttt{GWD}(\cdot, \cdot, \cdot)$ for structural variations.
% to consider both features and structures.
FGWD between two graphs $\OriGraph = (\nodes, \edges, \mat{X})$ and $\AugGraph = (\nodes^{+}, \edges^{+}, \mat{X}^{+})$ is as follows:
\begin{equation*}
	\texttt{FGWD}_{\alpha}(\OriGraph, \AugGraph) \coloneq \!\!\!\min_{\gmat{\pi}  \in \Pi(\gvect{\mu}, \gvect{\nu})} \!\! \alpha \cdot \texttt{WD}(\mat{X}, \mat{X}^{+}, \gmat{\pi}) + (1 - \alpha) \cdot
    % \texttt{GWD}(\mat{A}, \mat{A}^{+}, \gmat{\pi}),
    \texttt{GWD}(\edges, \edges^{+}, \gmat{\pi}),
\vspace{-2mm}
\end{equation*}
where $\alpha$ is a balancing hyperparameter,
$\gmat{\pi}$ is the coupling matrix~\cite{FGW} describing the probabilistic matching between two graphs,
$\Pi(\gvect{\mu}, \gvect{\nu})$
is the set of all coupling matrices that align $\gvect{\mu}$ and $\gvect{\nu}$,
$\gvect{\mu}\in\mathbb{R}^{n}$ and $\gvect{\nu}\in\mathbb{R}^{n^+}$ are the source and target distributions on $\nodes$ and $\nodes^{+}$, respectively.
% The problem is solved in $O(|\nodes|^2|\nodes^{+}|+|\nodes||\nodes^{+}|^{2})$ time~\cite{Peyre,TFGW} by reforming the problem as a quadratic optimization problem and solving it using block coordinate descent technique.
% This involves iteratively minimizing the loss to feature and structure spaces with respect to $\gmat{\pi}$ (refer to~\cite{TFGW} for details).
%
There are traditional metrics for measuring graph distance, but most of them focus only on structural differences, with a few heuristically considering features.
In contrast, FGWD 1) flexibly balances both differences by $\alpha$ within a single optimization,
2) is computationally tractable compared to NP-hardness of graph edit distance~\cite{GED}, and
3) provides precise distances by following the optimal transport setting. 
% \presecMargin

%\vspace{-1mm}
\section{Proposed Method}
% \postsecMargin
%\vspace{-1mm}
\label{sec:method}
%%%% Overview
We propose \textbf{\method}, a novel graph representation learning framework for accurate graph classification.
% \vspace{-2mm}
% \subsection{Overview}
% \label{sec:intro:overview}
%
The technical challenges in improving the performance of graph classification with augmented graphs are as follows:

\begin{itemize}[leftmargin=7.5mm]
    \item[\textbf{C1.}] {
        \textbf{Capturing augmentation-induced difference.}
        Previous approaches encourage the encoder to generate representations invariant to graph augmentation, thereby disregarding the difference between the augmented graph and its original.
        How can we train the model to capture such a difference?
    }
    \item[\textbf{C2.}] {
        \textbf{Measuring the difference gained from graph augmentation.}
        Augmentation techniques rely on the perturbation ratio to control the extent of changes.
        However, this ratio is not suitable for representing the actual difference due to randomness.
        How can we measure this difference?
    }
    \item[\textbf{C3.}] {
        \textbf{Training robust classifier.}
        Existing approaches produce augmentation-invariant representations, which limit the classifier's ability to handle the diversity from graph augmentations.
        How can we train a robust classifier to better generalize and fully utilize expressive graph representations?

    }
\end{itemize}

% ================ Figure 2: Proposed Method =========================
\begin{figure}[t]
	% \vspace{-3mm}
    \centering
    \includegraphics[width=0.95\linewidth]{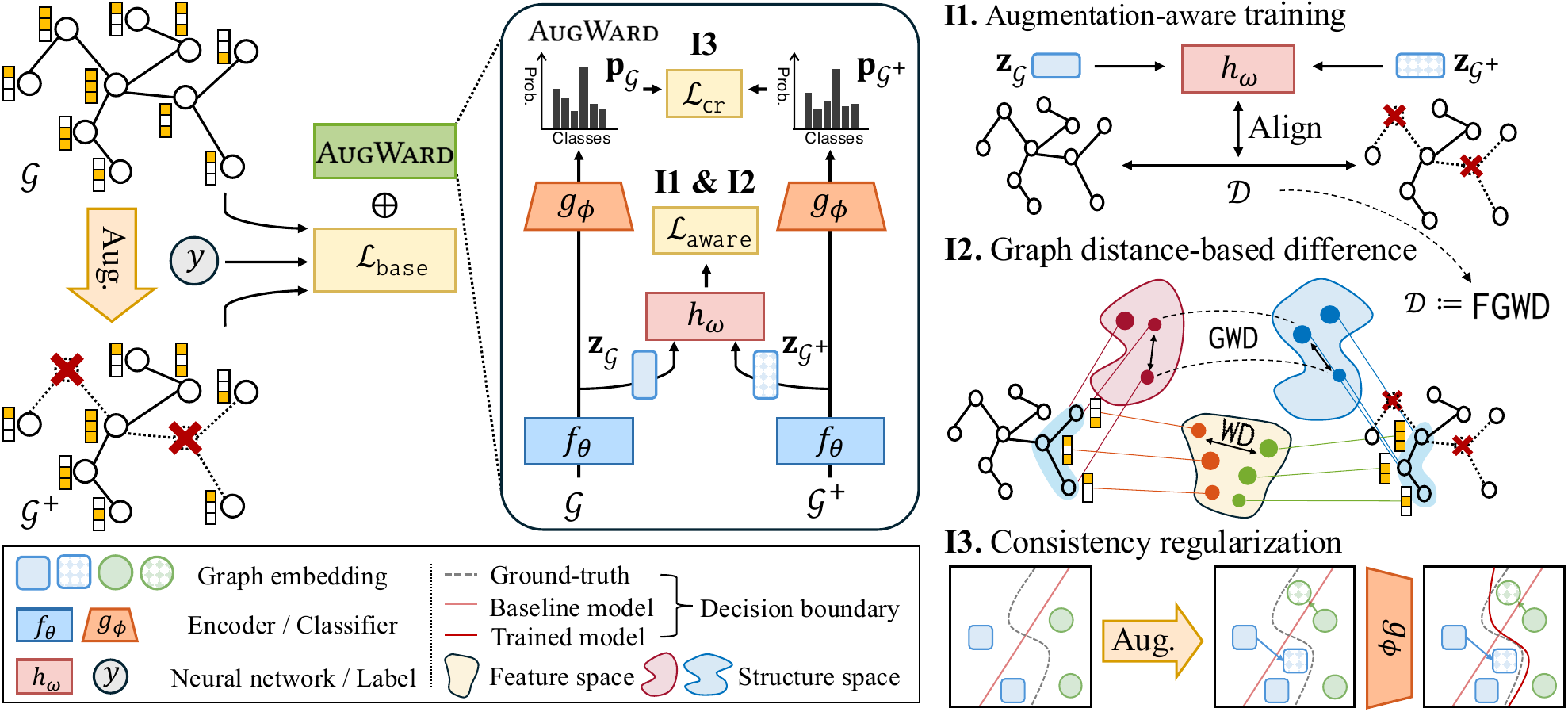}
%    \vspace{0.5mm}
    \caption{
        Overall process of \method.
        % when training a graph $\OriGraph$ with its augmented graph $\AugGraph$.
        \method incorporates augmentation-aware loss \(\Laware\) (\textbf{I1} \& \textbf{I2}) and consistency regularization loss \(\Lcr\) (\textbf{I3}).
        % , where
        % \(\Laware\) encourages a neural network $\neunet$ to output the Fused Gromov-Wasserstein distance $\texttt{FGWD}(\OriGraph, \AugGraph)$, while \(\Lcr\) aims to align the predictions from $\OriGraph$ and $\AugGraph$.
    }
  \label{fig:method}
  \vspace{1mm}
\end{figure}
% ================ END ===========================================

We propose the following ideas to deal with the challenges above:
\begin{itemize}[leftmargin=7.5mm]
    \item[\textbf{I1.}]{
        \textbf{Augmentation-aware training (Sec. \ref{subsec:I1}).}
        We propose a model-agnostic learning strategy that aligns the difference between the augmented graph and its original with that of their representations, enabling the training model to be aware of the augmentation-induced difference.
    }
    \item[\textbf{I2.}] {
        \textbf{Graph distance-based difference (Sec. \ref{subsec:I2}).}
        We measure the graph distance between the augmented graph and its original as their graph-level difference.
        Specifically, we exploit FGWD since it captures the variations in both features and structures induced by graph augmentation.
    }
    \item[\textbf{I3.}] {
        \textbf{Consistency regularization (Sec. \ref{subsec:I3}).}
        We jointly optimize consistency regularization term that aligns the predictions of an augmented graph with those of its original graph, fully harnessing expressive representations obtained through our augmentation-aware training.
    }
\end{itemize}

Figure~\ref{fig:method} depicts the overall procedure of \method.
% Our ideas are encapsulated as an augmentation-aware training framework with graph distance.
% An original graph $\OriGraph$ is augmented into $\AugGraph$ through a graph augmentation method $\augmenter$.
% %
% Afterward, $\OriGraph$ and $\AugGraph$ are encoded into graph representations $\OriZ$ and $\AugZ$, respectively, via a GNN encoder $\encoder$.
% To be aware of the augmentation-induced difference, \method additionally trains a neural network $\neunet$ with their representations, aiming to approximate the FGWD between the two graphs $\OriGraph$ and $\AugGraph$, where its objective function is denoted as $\Laware$.
% Each graph representation is fed into an MLP (Multi-Layer Perceptron) classifier $\classifier$, resulting in a vector of prediction scores, denoted respectively as $\OriPred$ and $\AugPred$.
% %
% Then, our \method aims to make them similar through the consistency regularization whose loss function is $\Lcr$.
Our framework jointly optimizes $\Laware$ and $\Lcr$ along with a baseline loss $\Lori$ (e.g., semi-supervised learning), where any technique is adoptable for $\augmenter$, $\encoder$, and $\classifier$.
% The details of the procedure are described in Algorithm~\ref{alg:method}.

%\presubsecMargin
%%%% Subsection 3.2. Augmentation-aware Training
\subsection{Augmentation-aware Training}
\label{subsec:I1}
\begin{figure}[t]
    \centering
    \begin{minipage}[t]{0.45\linewidth}
         \centering
         \includegraphics[width=\linewidth]{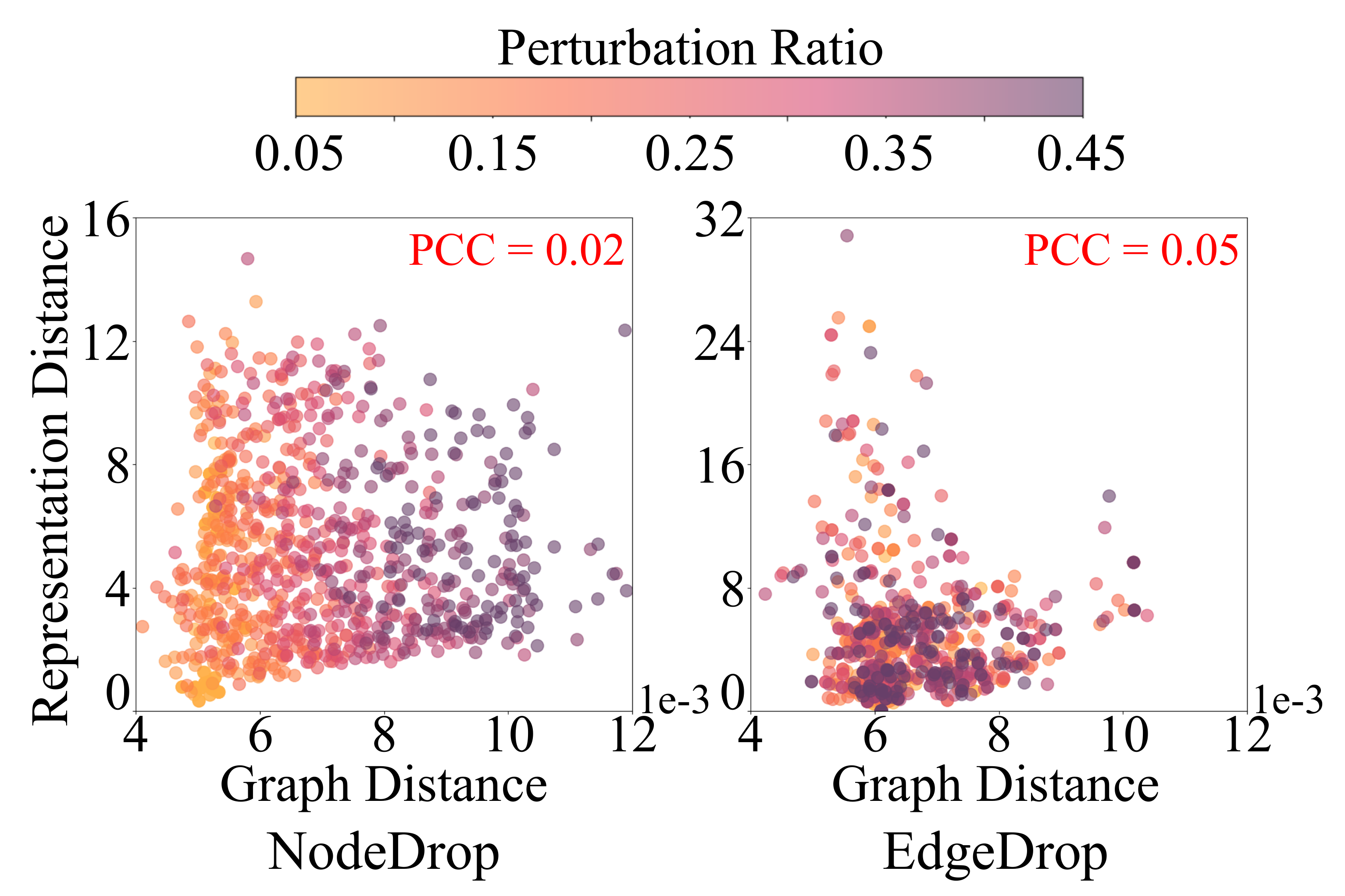}
%         \vspace{0.5mm}
         \caption{The representation-level and graph-level differences between $\OriGraph$ and $\AugGraph$ show no correlation.
         See Section~\ref{subsec:I1} for details.}
         \label{fig:nonlinear}
     \end{minipage}
     \hspace{0.01\linewidth}
    \begin{minipage}[t]{0.45\linewidth}
         \centering
         \includegraphics[width=\linewidth]{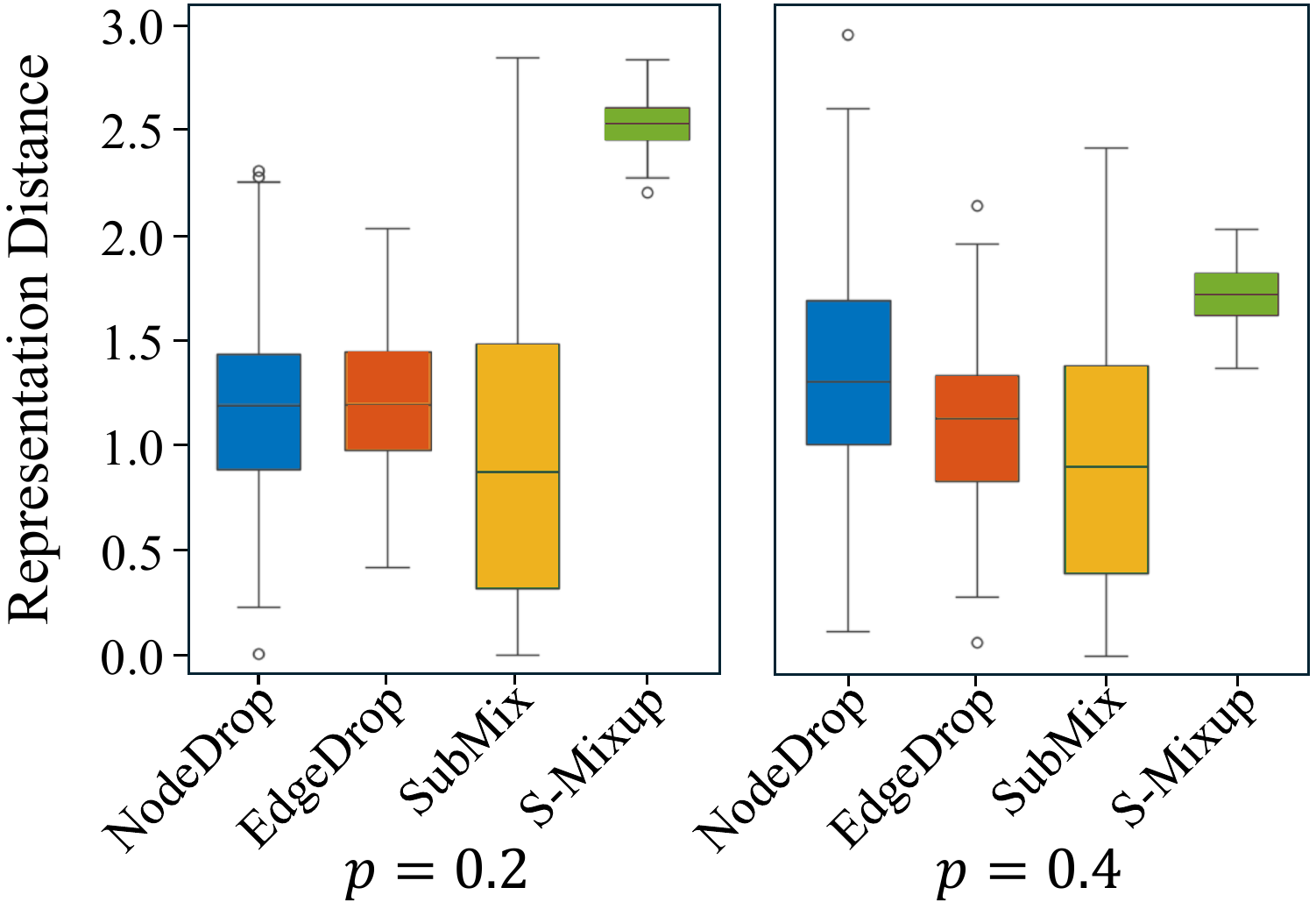}
%         \vspace{0.5mm}
         \caption{
         % The representation-level distance distribution of 100 augmented graphs at a fixed perturbation ratio $p$.
         Graphs with the same perturbation ratio $p$ demonstrate notable diversity across various augmentations.
         See Section~\ref{subsec:I2} for details.}
         \label{fig:randomness}
     \end{minipage}
    \label{fig:3+4}
%    \vspace{3mm}
\end{figure}

%\postsubsecMargin
%
\smallsection{Observation}
We first present an empirical observation that explains why the existing approaches fail to reflect the difference induced by augmentation in the representations between the augmented graph $\AugGraph$ and its original $\OriGraph$.
We assume that ideal representations are able to distinguish graph-level differences, and thus their representations should differ in proportion to how much the augmented graph deviates from the original graph, i.e., there should be a positive correlation between them.
To check this, we begin by applying augmentations of varying intensities to a graph $\OriGraph$ sampled from the \proteins~\cite{TUDataset} dataset.
For each perturbation ratio $p \in \{0.05, 0.1, \cdots, 0.45\}$, we generate $100$ augmented graphs $\AugGraph$.
After training a GIN encoder, we measure the Euclidean distance $\lVert \OriZ - \AugZ \rVert_{2}^{2}$ of their representations and the actual graph distance (in FGWD) between $\OriGraph$ and $\AugGraph$.
As shown in Figure~\ref{fig:nonlinear}, Pearson Correlation Coefficients (PCC) are almost zero, indicating that there is no strong correlation between representation-level and graph-level differences.
% This pattern is commonly seen with various augmentations (refer to Supplementary Materials~\cite{supp} for other augmentations).
This clearly verifies the loss of augmentation-induced differences in existing representation learning methods.

\smallsection{Solution}
Motivated by this observation, we aim to encourage the encoder $\encoder$ to align the representation-level difference with the graph-level difference.
For this purpose, we train a neural network $\neunet$ with parameters $\omega$ using the graph representations $\OriZ$ and $\AugZ$ to capture the difference between the original and augmented graphs $\OriGraph$ and $\AugGraph$.
This is achieved by optimizing $\Laware$:
\begin{equation}
	\label{eq:eq5}
	\Laware\bigl(\OriGraph, \AugGraph\bigr) \coloneq \bigl\lVert \: \neunet \bigl(\OriZ, \AugZ\bigr) - \mathcal{D}\bigl(\OriGraph, \AugGraph\bigr) \: \bigr\rVert^2_{2},
\end{equation}
where $\mathcal{D}(\OriGraph, \AugGraph)$ denotes the difference between the graphs, and we employ a fully connected layer $\neunet$ with the concatenation of $\OriZ$ and $\AugZ$ as input for $\neunet$.
The loss injects the graph-level difference $\mathcal{D}$ into the relationship between the representations, captured by the neural network $\neunet$, so that the difference information is encoded into the representations.
Note that there would be various options for $\mathcal{D}$, and any choice would be adopted within the loss function.
Which is suitable for measuring the difference, especially in the context of graph classification?
We present a detailed discussion on our design choice of $\mathcal{D}$ in Section~\ref{subsec:I2}.

% ================ Figure 3: Difference between representation and graphs ==================
% ================ END ===========================================

%\vspace{-1mm}
% \presubsecMargin
%%%% Subsection 3.3. Graph Distance Computation
\subsection{Graph Distance-based Difference}
\label{subsec:I2}
%\postsubsecMargin
%
\smallsection{Observation}
A na\"ive answer for $\mathcal{D}(\OriGraph, \AugGraph)$ is the perturbation ratio $p$, as it indicates the amount of change from the original graph through graph augmentation $\augmenter$. 
However, this approach fails to precisely represent the difference between $\OriGraph$ and $\AugGraph$, because generated graphs at a fixed perturbation ratio $p$ exhibit significant variations due to inherent randomness.
To investigate this, we generate $100$ augmented graphs $\AugGraph$ from a sampled graph $\OriGraph$ from the \proteins dataset for each augmentation method $\augmenter$ with $p$ fixed at $0.2$ and $0.4$, respectively.
Then, we measure the Euclidean distances $\lVert \OriZ - \AugZ \rVert_{2}^{2}$
% , where $\OriZ$ and $\AugZ$ are the representation of graphs $\OriGraph$ and $\AugGraph$, respectively
(i.e. $\OriZ = \encoder(\OriGraph)$, $\AugZ = \encoder(\AugGraph)$).
Figure~\ref{fig:randomness} illustrates the distributions of the representation-level distances for each $p$ and augmentation method.
These distributions show significant variance and inconsistency across different augmentation types and ratios.
Hence, the perturbation ratio $p$ is inappropriate to represent the difference $\mathcal{D}$.

% ================ Figure 5: Examples ==================
\begin{figure}[t]
	\centering
	\includegraphics[width=0.85\linewidth]{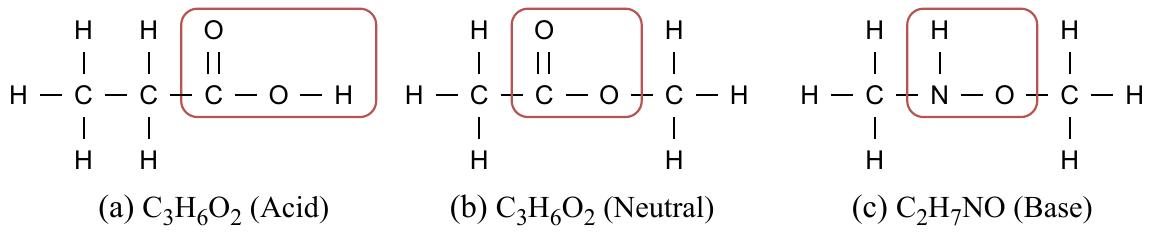}
	\vspace{-1mm}
	\caption{Graphs of three compounds. (a) and (b) share graph features, while (b) and (c) exhibit identical graph structures.
	Functional groups are marked in red.}
	\label{fig:examples}
%	\vspace{1mm}
\end{figure}
%
%% ================ END ===========================================

%\vspace{1mm}
\smallsection{Solution}
Our idea is to employ a graph distance metric to explicitly measure the difference between $\OriGraph$ and $\AugGraph$.
Specifically, we exploit Fused Gromov-Wasserstein Dimstance (FGWD) for this purpose.
The main reason is as follows.
In the graph classification task, each graph $\Graph$ includes its topological structure, defined by $\nodes$ and $\edges$, along with node features $\mat{X}$.
Graph augmentation alters either structure or features, which are effectively captured by FGWD.
Figure~\ref{fig:examples} shows examples of chemical compound graphs, showing why considering both structure and features is important.
In these graphs, (a) and (b) share identical features, while (b) and (c) exhibit the same structure.
The difference in feature or structure,
%even if it is just one aspect,
leads to a distinction in chemical types.
Therefore, it is desirable to consider differences in both structure and features.

Although there are traditional graph distance metrics, most consider only structural information or rely on heuristics to consider features.
In contrast, FGWD effectively captures both aspects, showing advantages as discussed in Section~\ref{sec:prelim}.
Thus, we measure FGWD as the difference $\mathcal{D}$ between $\OriGraph$ and  $\AugGraph$.
The loss function of our augmentation-aware training is represented as follows:
\vspace{-1mm}
\begin{equation}
	\label{eq:aware:fgw}
	\Laware\bigl(\OriGraph, \AugGraph\bigr) \coloneq \bigl\lVert \: \neunet \bigl(\OriZ, \AugZ\bigr) - \texttt{FGWD}_{\alpha}\bigl(\OriGraph, \AugGraph\bigr) \: \bigr\rVert^2_{2},
	\vspace{-1mm}
\end{equation}
where $\alpha$ is the hyperparameter of FGWD.
% In the computation of FGWD,
We set $\gvect{\mu} = \vect{1}_{n}/n$ and $\gvect{\nu} = \vect{1}_{n^+}/n^+$ following~\cite{FGWMixup}, where $n = |\nodes|$, $n^{+} = |\nodes^{+}|$, and $\vect{1}_{n}$ is an all-one vector of size $n$.

% \vspace{-1mm}
%\presubsecMargin
%%%% Subsection 3.4. Consistency Regularization
\subsection{Consistency Regularization}
\label{subsec:I3}
\vspace{-1mm}
%\postsubsecMargin
%
\smallsection{Motivation}
%The existing approaches (e.g., contrastive learning) mainly focus on producing augmentation-invariant representations, without paying attention to the classifier, as they expect similar representations to be helpful for consistent prediction.
%However, as discussed in Section~\ref{sec:intro}, these approaches limit the expressiveness of representations, preventing the classifier from being adequately trained to handle the diversity introduced by graph augmentations.
%
%
Given distinguishable representations, it is important to train the classifier robustly for better generalization.
In node classification, GRAND~\cite{GRAND} points out that matching predictions from different representations for the same label improves the model's generalization behavior.
Inspired by this, we design a loss for consistency regularization for graph classification, so that the classifier
$\classifier$ fully utilizes the expressive representations by augmentation-aware training.

\smallsection{Solution}
As shown in Figure~\ref{fig:method}, the classifier $\classifier$ yields $\OriPred$ and $\AugPred$, vectors of prediction scores, from $\OriZ$ and $\AugZ$. %, i.e., $\OriPred \leftarrow \classifier(\OriZ)$, respectively.
The idea of the consistency regularization is to match the two predictions $\OriPred$ and $\AugPred$, with the loss represented as follows:
\vspace{-2mm}
\begin{align}
    \begin{split}
        \label{eq:cr}
        \Lcr\bigl(\OriGraph, \AugGraph\bigr) \coloneq H\bigl(\OriPred, \AugPred\bigr)
        = -\sum_{i=1}^{|C|} P(y = i | \OriGraph) \cdot \log{P(y = i | \AugGraph)},
    \end{split}
    \vspace{-4mm}
\end{align}
where $H$ denotes the cross-entropy, and the $i$-th entry of $\OriPred$ indicates $P(y = i | \OriGraph)$ (similar for $\AugGraph$).
%
% Implication & result
In other words, it regulates the classifier to make consistent predictions for both $\OriGraph$ and $\AugGraph$, aiming to improve robustness and accuracy.

%\presubsecMargin
%\vspace{-1mm}
%%%% Subsection 3.5. Final Loss Function
\subsection{Final Loss Function}
%\postsubsecMargin
\vspace{-1mm}
We sum up all the loss functions for \method as follows:
\vspace{-1mm}
\begin{equation}
	\label{eq:loss:augward}
    \Laugward(\OriGraph, \AugGraph) \coloneq \LambdaAware\cdot\Laware(\OriGraph, \AugGraph) + \LambdaCR\cdot\Lcr(\OriGraph, \AugGraph),
    \vspace{-1mm}
\end{equation}
where $\LambdaAware$ and $\LambdaCR$ are hyperparameters that control the strength of their respective losses.
Our framework supports various baseline learning paradigms including supervised, semi-supervised, and transfer learning for graph classification.
This is achieved by jointly optimizing $\Lori$ and $\Laugward$ as follows:
\begin{equation}
    \label{eq:eq10}
    \mathcal{L}(\OriGraph, \AugGraph, y) \coloneq \Lori(\OriGraph, \AugGraph, y) + \Laugward(\OriGraph, \AugGraph),
%    \vspace{-1mm}
\end{equation}
where $\Lori(\OriGraph, \AugGraph, y)$ is the baseline loss with the ground-truth label $y$ (e.g., the sum of cross-entropy losses for $\OriGraph$ and $\AugGraph$ in supervised learning).

\section{Experiments}
%\postsecMargin
\label{sec:experiments}
% ================ Table 1: Supervised Graph Classification ============
\def\arraystretch{1.0}
\begin{table*}[t]
	\vspace{-2mm}
	\centering
	\caption{
	Accuracy of supervised graph classification for various graph augmentation methods, where "Imp." indicates the percentage point improvement of the average accuracy when \method is applied.
	\method consistently enhances the performance with those augmentation methods on various datasets.
    \vspace{1mm}
    }
	\resizebox{0.95\linewidth}{!}{
        \setlength{\tabcolsep}{4.5pt}
		\begin{tabular}{lcccccccccc|c|c}
			\hline
			\toprule
			\textbf{Models} & \textbf{\dd} &  \textbf{\enzymesShort}$^*$&\textbf{\imdbbShort}$^*$ & \textbf{\imdbmShort}$^*$ & \textbf{\nciOne} &\textbf{\nciNine}&\textbf{\ptcmrShort}$^* $&\textbf{\proteinsShort}$^*$& \textbf{\redditbShort}$^*$ & \textbf{\twitterShort}$^*$ & \textbf{Avg.} & \textbf{Imp.}\\
			\midrule
			NodeDrop~\cite{GraphCL} & 75.42 & 27.67 & 56.40 & 44.53 & 67.15 & 65.04 & 67.24 & 70.95 & 76.30 & 65.14 & 61.58 & - \\
			\textbf{+ \method} & 76.69 & 28.67 & 58.40 & 45.87 & 67.79 & 65.52 & 76.44 & 72.76 & 78.40 & 65.30 & \textbf{63.58} & \textbf{+2.00} \\
			\midrule
			EdgeDrop~\cite{GraphCL}& 71.39 & 26.67 & 55.20 & 46.13 & 66.13 & 64.79 & 68.94 & 67.39 & 77.10 & 65.16 & 60.89 & - \\
			\textbf{+ \method} & 75.84 & 29.00 & 60.60 & 46.93 & 67.10 & 65.08 & 75.29 & 67.20 & 77.50 & 65.61 & \textbf{63.02} & \textbf{+2.13} \\
			\midrule
			AttrMask~\cite{GraphCL}& 69.15 & 26.67 & 56.60 & 46.93 & 65.84 & 64.60 & 68.37 & 67.44 & 74.50 & 65.10 & 60.52 & - \\
			\textbf{+ \method} & 72.54 & 29.33 & 58.80 & 48.40 & 67.06 & 64.94 & 73.01 & 67.92 & 78.30 & 65.50 & \textbf{62.58} & \textbf{+2.06} \\
			\midrule
			Subgraph~\cite{GraphCL} & 71.02 & 25.00 & 56.40 & 46.53 & 64.40 & 64.67 & 69.56 & 64.34 & 75.30 & 64.89 & 60.21 & - \\
			\textbf{+ \method} & 72.71 & 25.33 & 60.40 & 45.20 & 65.57 & 65.11 & 72.39 & 69.35 & 75.10 & 65.08 & \textbf{61.62} & \textbf{+1.41} \\
			\midrule
			GraphAug~\cite{GraphAug}& 71.53 & 27.67 & 57.40 & 45.92 & 64.96 & 61.26 & 68.35 & 69.52 & 77.20 & 65.05 & 60.89 & - \\
			\textbf{+ \method} & 73.90 & 28.00 & 60.60 & 46.77 & 66.96 & 62.28 & 70.10 & 70.54 & 78.20 & 65.21 & \textbf{62.26} & \textbf{+1.37} \\
			\midrule
			SubMix~\cite{NodeSam}& 79.83 & 27.00 & 59.40 & 45.11 & 64.70 & 63.83 & 70.67 & 68.63 & 76.40 & 64.67 & 62.02 & - \\
			\textbf{+ \method} & 80.00 & 27.67 & 60.80 & 45.07 & 65.06 & 64.02 & 71.28 & 68.83 & 79.00 & 65.37 & \textbf{62.71} & \textbf{+0.69} \\
			\midrule
			S-Mixup~\cite{S-Mixup} & 79.29 & 27.33 & 58.60 & 46.67 & 65.11 & 64.26 & 68.35 & 62.54 & 77.60 & 64.96 & 61.47 & - \\
			\textbf{+ \method} & 79.49 & 26.33 & 59.80 & 46.80 & 65.50 & 64.60 & 70.67 & 65.69 & 76.80 & 65.34 & \textbf{62.10}& \textbf{+0.63}\\
			\bottomrule
			\hline
			\multicolumn{12}{l}{\scriptsize * \enzymesShort: \enzymes, \imdbbShort: \imdbb, \imdbmShort: \imdbm, \ptcmrShort: \ptcmr, \proteinsShort: \proteins, \redditbShort: \redditb, \twitterShort: \twitter}
		\end{tabular}
	}
	\label{tab:q1}
%	\vspace{2mm}
\end{table*}

% ================ END ===========================================

We perform experiments to answer the following questions:
\begin{itemize}[leftmargin=7.5mm]
    \item[\textbf{Q1.}] {
        \textbf{Accuracy in supervised graph classification (Section \ref{subsec:q1}).}
    	Does \method improve the graph classification accuracy of supervised classification methods utilizing graph augmentation techniques?
    }
    \item[\textbf{Q2.}] {
        \textbf{Accuracy in semi-supervised graph classification (Section \ref{subsec:q2}).}
        Does \method improve the graph classification accuracy of semi-supervised classification methods utilizing graph augmentation techniques?
    }
    \item[\textbf{Q3.}] {
        \textbf{Representation transferability
        (Section \ref{subsec:q3}).}
        How accurate is \method in transfer learning? Does it enhance representation transferability?
    }
     \item[\textbf{Q4.}] {
        \textbf{Runtime analysis (Section \ref{subsec:q4}).}
        How significant is the additional computational overhead caused by \method?
    }
    \item[\textbf{Q5.}] {
        \textbf{Ablation study (Section \ref{subsec:q5}).}
        Are all the components of \method effective in improving model performance?
        % How do the hyperparameters of \method affect its performance?
    }
\end{itemize}
%
% \presubsecMargin
%%%% Subsection 4.1. Experimental Setup
% \subsection{Experimental Setup}
% \label{subsec:setup}
\smallsection{Experimental Setup}
% \postsubsecMargin
%
%
%
We evaluate \method in various settings, including supervised, semi-supervised, and transfer learning.
Following~\cite{GIN,NodeSam}, we use ten benchmark datasets~\cite{TUDataset} for supervised and semi-supervised learning, where the label rate is 10\% for the semi-supervised setting.
For transfer learning, we evaluate \method with \texttt{ZINC15}~\cite{Zinc15} as the source and eight downstream datasets from MoleculeNet~\cite{MoleculeNet} as the target~\cite{ContextPred}. 
We set the learning rate to $0.01$ for the Adam optimizer and train a four-layered GIN for 100 epochs.
We conduct a grid search of hyperparameters: batch size $\in$ \{32, 128\}, dropout $\in$ \{0, 0.5\}, $p \in$ \{0.05, 0.1, 0.15, 0.2\}, $\alpha \in$ \{0.05, 0.5, 0.95\}, $\LambdaAware \in$ \{5, 10, 25, 50, 75, 100\}, and $\LambdaCR \in$ \{0, 0.1, 1, 10, 100\}.
All experiments are done on a single RTX 3090 GPU. %
% We implement \method with PyTorch~\cite{PyTorch} and PyTorch Geometric~\cite{PyTorch_Geometric} in Python.

%%%% Subsection 4.2. Supervised Graph Classification (Q1)
\subsection{Supervised Graph Classification (Q1)}
 \label{subsec:q1}
We examine the effect of \method for various augmentations on supervised graph classification.
We train with each augmentation method as baseline and compare with \method.
Table~\ref{tab:q1} shows the results on various datasets, and reports the average accuracy ("Average") and the percentage point improvement ("Imp.").
%
% ================ Table 2: Semi-supervised Accuracy =========================
\begin{table*}[t]
	\vspace{-2mm}
	\centering
	\caption{
	Accuracy [\%] of semi-supervised graph classification using NodeDrop as graph augmentation within each baseline, where only 10\% labeled graphs are available for training.
	\method provides improved performance in average accuracy, verifying its effectiveness also in the semi-supervised setting.
	%Comparative analysis of semi-supervised graph classification (10\% labeled data). We utilize NodeDrop as the graph augmentation. Note the consistent performance improvement facilitated by \method in a semi-supervised learning context.
%    \vspace{1mm}
	}
	\def\arraystretch{0.9}
	\resizebox{0.95\linewidth}{!}{
        \setlength{\tabcolsep}{5pt}
		\begin{tabular}{lcccccccccc|c|c}
			\hline
			\toprule
			\textbf{Models} & \textbf{\dd} &  \textbf{\enzymesShort}$^*$&\textbf{\imdbbShort}$^*$ & \textbf{\imdbmShort}$^*$ & \textbf{\nciOne} & \textbf{\nciNine}&\textbf{\ptcmrShort}$^* $&\textbf{\proteinsShort}$^*$& \textbf{\redditbShort}$^*$ & \textbf{\twitterShort}$^*$ & \textbf{Avg.} & \textbf{Imp.}\\
			\midrule
			InfoGraph \cite{InfoGraph}& 63.64& 20.67 & 50.20 & 37.33 & 61.68 & 61.91 & 64.91 & 65.39 & 66.20 & 55.29 & 54.72 & - \\
			\textbf{+ \method} & 63.98 & 22.67 & 54.60 & 38.79 & 62.89 & 62.44 & 66.08 & 66.29 & 67.50 & 57.18 & \textbf{56.24} & \textbf{+1.52} \\\midrule
			GraphCL \cite{GraphCL} & 63.39 & 21.00 & 53.80 & 39.47 & 61.92 & 61.19 & 66.66 & 64.37 & 67.10 & 56.71 & 55.56 & - \\
			\textbf{+ \method} & 64.18 & 23.67 & 55.20 & 39.73 & 63.50 & 62.83 & 67.24 & 65.93 & 68.20 & 57.09 & \textbf{56.76} &	\textbf{+1.20}\\\midrule
			CuCo \cite{CuCo} & 63.81 & 23.33 & 52.40 & 39.27 & 61.97 & 60.40 & 65.61 & 65.57 & 67.40 & 57.11 & 55.69 & -\\
			\textbf{+ \method} & 64.66 & 24.00 & 54.20 & 39.80 & 62.17 & 61.21 & 68.78 & 67.39 & 68.00 & 57.28 & \textbf{56.75} & \textbf{+1.06} \\\midrule
			GCL-SPAN \cite{GCL-SPAN} & 63.22 & 23.67 & 53.20 & 40.20 & 62.96 & 61.46 & 65.64 & 66.84 & 67.50 & 57.73 & 56.24 & -\\
			\textbf{+ \method} & 64.26 & 24.33 & 55.80 & 41.26 & 64.47 & 62.05 & 67.39 & 67.80 & 68.30 & 57.92 & \textbf{57.36} & \textbf{+1.12} \\
			\bottomrule
			\hline
			\multicolumn{13}{l}{\scriptsize * \enzymesShort: \enzymes, \imdbbShort: \imdbb, \imdbmShort: \imdbm, \ptcmrShort: \ptcmr, \proteinsShort: \proteins, \redditbShort: \redditb, \twitterShort: \twitter}
		\end{tabular}
	}
	\label{tab:q2}
	\vspace{-5mm}
\end{table*}
% ================ END ===========================================
%
% As shown in the table,
\method consistently improves accuracy for all tested augmentation methods across most datasets, with increases in average accuracy of up to 2.13\%p.
This indicates that \method effectively improves learning performance, while being easily integrated.

%%%% Subsection 4.3. Semi-supervised Setting (Q2)
\subsection{Semi-supervised Graph Classification (Q2)}
\label{subsec:q2}
We evaluate the effectiveness of \method for Semi-Supervised Learning (SSL) methods on graph classification, where labels are available for 10\% of the graphs.
We choose NodeDrop~\cite{GraphCL} as the default augmentation for each SSL method.
Table~\ref{tab:q2} shows \method enhances the SSL models across various datasets, achieving up to 1.52\%p increase in average accuracy.
The result verifies that our strategy is beneficial also for SSL, as well as supervised learning.

% ================ Table 3: Graph Transfer Learning Accuracy =========================
\begin{table*}[t]
	\def\arraystretch{0.9}
    \vspace{-2mm}
	\centering
	\caption{
		ROC-AUC of transfer learning experiments, pretrained on \zinc~\cite{Zinc15} and fine-tuned on MoleculeNet~\cite{MoleculeNet} datasets.
		\method improves existing models for transfer learning, offering more expressive graph representations.
%        \vspace{1mm}
	}
        \resizebox{0.95\linewidth}{!}{
		\setlength{\tabcolsep}{6.3pt}
		\footnotesize
		\begin{tabular}{lcccccccc|c|c}
			\hline
			\toprule
			\textbf{Models} &
			\textbf{\texttt{BACE}} & \textbf{\texttt{BBBP}} & \textbf{\texttt{ClinTox}} & \textbf{\texttt{HIV}} & \textbf{\texttt{MUV}} & \textbf{\texttt{Tox21}} & \textbf{\texttt{ToxCast}} & \textbf{\texttt{SIDER}} & \textbf{Avg.} & \textbf{Imp.} \\
			\midrule
			ContextPred \cite{ContextPred}& 80.80 & 71.76 & 70.22 & 77.64 & 76.14 & 75.56 & 63.00 & 61.43 & 72.07 & - \\
			\textbf{+ \method} & 83.32 & 72.85 & 70.60 & 79.35 & 80.80 & 75.81 & 63.94 & 62.82 & \textbf{73.69} & \textbf{+1.62} \\\midrule
			GraphCL \cite{GraphCL} & 73.15 & 70.33 & 73.80 & 80.08 & 69.30 & 74.44 & 62.32 & 60.48 & 70.49 & - \\
			\textbf{+ \method} & 77.24 & 72.78 & 82.41 & 81.42 & 77.67 & 75.69 & 63.46 & 61.66 & \textbf{74.04} & \textbf{+3.55} \\\midrule
			MGSSL \cite{MGSSL}& 80.83 & 72.76 & 76.74 & 73.30 & 72.35 & 74.87 & 62.02 & 56.07 & 71.12 & - \\
			\textbf{+ \method} & 85.41 & 74.26 & 82.15 & 77.95 & 77.20 & 76.09 & 63.81 & 61.77 & \textbf{74.83} & \textbf{+3.71}  \\\midrule
			GraphMAE \cite{GraphMAE} & 81.30 & 72.04 & 82.82 & 77.15 & 72.21 & 75.33 & 64.07 & 61.07 & 73.25 & - \\
			\textbf{+ \method} & 83.50 & 73.19 & 84.88 & 78.63 & 81.86 & 75.76 & 64.08 & 61.62 & \textbf{75.44} & \textbf{+2.19} \\
			\bottomrule
			\hline
		\end{tabular}
        }
	\label{tab:q3}
	\vspace{1mm}
\end{table*}
% ================ END ===========================================
%
%%%% Subsection 4.4. Representation Transferability (Q3)
\subsection{Representation Transferability (Q3)}
\label{subsec:q3}
We investigate how \method improves the transferability of graph representations in transfer learning models.
For this, we pretrain a model on the \zinc~\cite{Zinc15} dataset and fine-tune it on eight downstream datasets from MoleculeNet~\cite{MoleculeNet}.
% Similarly, we repeat this process with each model trained using our \method.
As shown in Table~\ref{tab:q3}, \method consistently improves the performance of different transfer learning models for most downstream tasks in terms of ROC-AUC, specifically up to 3.71\%p higher average accuracy.
This indicates that \method produces enhanced representations for transfer learning.
%This indicates that \method produces representations with enhanced expressiveness that are effective in transfer learning.

%\vspace{-1mm}

%%%% Subsection 4.5. Runtime Analysis (Q4)
\subsection{Runtime Analysis (Q4)}
\label{subsec:q4}
We analyze the computational overhead of \method.
For this, we measure the training time of \method
% Algorithm~\ref{alg:method}
, which consists of 1) baseline supervised learning with augmentation, 2) computing FGWD, and 3) other remaining parts.
Figure~\ref{fig:runtime} shows the proportion of each component over the total training time for five datasets of various sizes.
Note that the overhead from FGWD is marginal, i.e., computing FGWD takes 4.89\% of the total time in average.
%This is because the number of nodes for each graph in real-world datasets is small.
%, which does not burden the FGWD calculation.
Thus, \method achieves a favorable trade-off between time and accuracy.
% , improving adopted models with little sacrifice of training time.

% ================ END ===========================================
%

%%%% Subsection 4.6. Ablation Study (Q5)
\subsection{Ablation Study (Q5)}
\label{subsec:q5}
\begin{figure}[t]
    \centering
    \begin{minipage}[t]{0.42\linewidth}
        \centering
        \includegraphics[width=\linewidth]{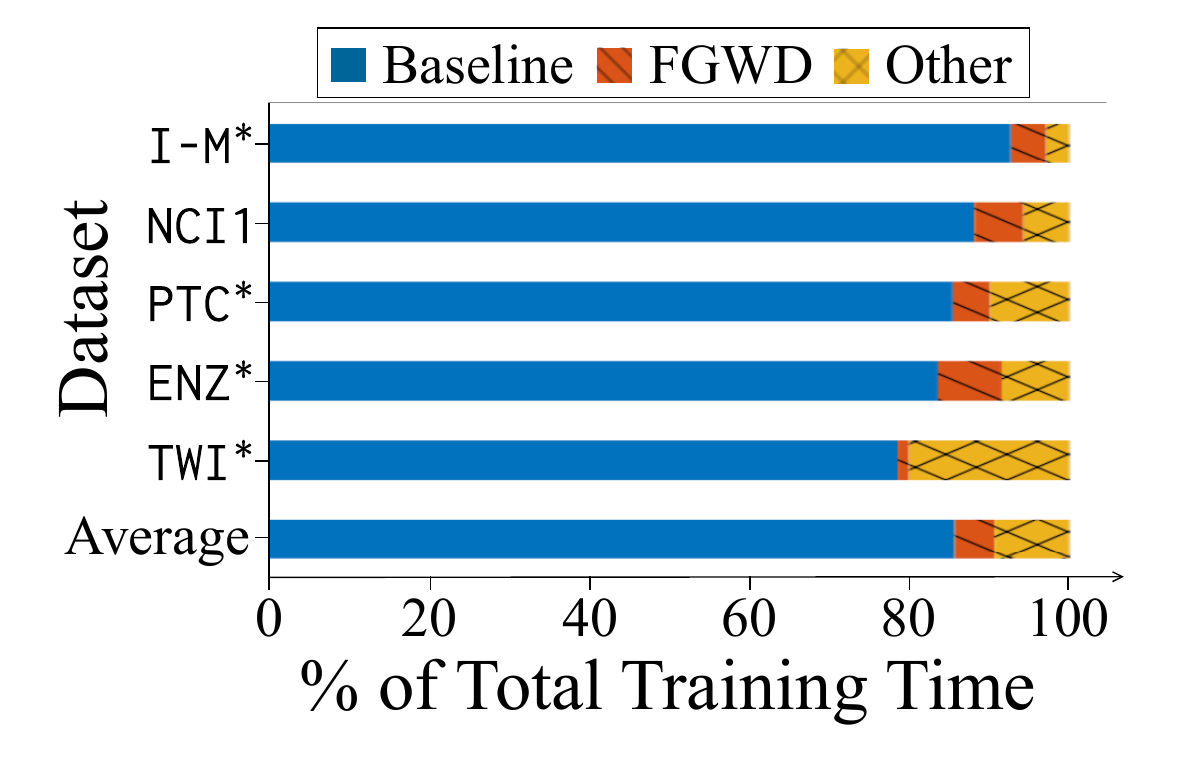}
%		 \vspace{1mm}
        \caption{
		  Proportion of running time in \method components.
		  %FGWD computation does not result in major overhead.
%		Percentage of training time for three components of \method.
%		Additional overhead of \method due to FGWD computation is marginal.
		\label{fig:runtime}
        }
     \end{minipage}
     \hspace{0.01\linewidth}
    \begin{minipage}[t]{0.50\linewidth}
        \vspace{-36mm}
        \setlength{\tabcolsep}{7.8pt}
        \small
        \centering
        \captionsetup{type=table}
        \caption{
        Ablation study for our proposed ideas in \method.
        All ideas of \method enhance the performance.
%        \vspace{1mm}
        }
        \resizebox{\linewidth}{!}{
            \begin{tabular}{l|l|c|c}
                \hline
                \toprule
                \textbf{Variants} & \textbf{Ideas} & \textbf{Avg.} & \textbf{Imp.}\\
                \midrule
                A: GIN + NodeDrop & Existing & 61.58 & - \\
                \midrule
                A + $p$ & I1 & 61.90 & +0.32 \\
                A + NFs & I1 & 62.50 & +0.92 \\
                A + AMs & I1 & 62.55 & +0.98 \\
                A + Edge Jaccard & I1  & 62.53  & +0.95\\
                \midrule
%                A + NFs and AMs & I1+I2 & 62.63 & +1.05 \\
                A + FGWD & I1+I2 & 63.04  & +1.46 \\
                \midrule
                A + \textbf{\method} & I1+I2+I3 &  \textbf{63.58} & \textbf{+2.00} \\
                \bottomrule
                \hline
            \end{tabular}
        }
        \label{tab:ablation}
    \end{minipage}
    \label{fig6+tab4}
%    \vspace{2mm}
\end{figure}

To evaluate the effectiveness of our proposed ideas, we conduct an ablation study by incorporating each idea incrementally.
Starting with GIN and NodeDrop as the baseline, we examine four variants for augmentation-aware training (Idea 1), where the augmentation-induced difference $\mathcal{D}$ formulated as:
(1) perturbation ratio $p$,
(2) difference of Node Features (NFs),
(3) difference of Adjacency Matrices (AMs),
and (4) edge Jaccard similarity.
We further integrate graph distance-based difference (Idea 2) using FGWD and finally apply consistency regularization (Idea 3) to demonstrate the cumulative benefit of all three ideas.
Table~\ref{tab:ablation} reports the average classification accuracy across ten datasets where we have four observations.
First, considering the difference is beneficial, even though they are simple heuristics.
Second, $p$ is less effective than other variants.
Third, selecting FGWD as $\mathcal{D}$ shows the best performance among different metrics.
Last, the highest accuracy is achieved through the joint application of all the three ideas.
Overall, all our proposed ideas contribute to the enhanced performance.
%

% ================ Table 7 : Ablation Study ==================
\section{Related Work}
%\postsecMargin
\vspace{-1mm}
\label{sec:related}
% In this section, we review the previous studies on graph classification.
% representation learning for graph classification.
%
\vspace{-1mm}
\smallsection{Graph representation learning}
Learning graph representation~\cite{PULL,SIDE} plays a crucial role in classifying graphs, enabling models to make predictions on graphs~\cite{PULearning,GRAB}.
Previous studies have shown that GNN-based methods~\cite{GraphMAE,GCL-SPAN} perform better than similarity-based approaches such as graph kernels~\cite{ARK}.
These methods employ GNNs to capture higher-order structures through multi-layered message-passing, and yield representations via pooling.
For better generalization, recent studies exploit augmentation and further train with advanced strategies such as mutual information maximization~\cite{InfoGraph} and contrastive learning~\cite{CuCo,GraphCL}.

\smallsection{Learning graphs with augmentation}
Graph augmentation produces a new set of graphs from an original set, ensuring the new graphs share similar characteristics with their originals.
There are three categories of graph augmentation: structure-oriented, feature-oriented, and mixup-based methods.
Structure-oriented techniques~\cite{GraphAug,NodeAug} modify graph structures by randomly dropping nodes and edges or rewiring nodes.
Feature-oriented approaches~\cite{GraphCL} alter node or edge features by randomly masking or shuffling them.
Mixup-based methods~\cite{FGWMixup,NodeSam} create new graphs by interpolating between pairs of graphs.
However, these methods focus on augmentation-invariant representation, thus fail to capture the diversity between graphs in their representations.
\method effectively captures augmentation-induced differences and improves model performance. 
%\presecMargin

\section{Conclusion}
%\postsecMargin
\label{sec:conclusion}
We propose \method, a novel graph representation learning framework that considers augmentation-induced differences for accurate graph classification.
Our main idea is augmentation-aware training with graph distance and consistency regularization for improving both the quality of the graph representations and prediction accuracy.
Experimental results show that \method is robust and adaptable, improving performance in diverse settings such as augmentation, mixup, supervised, semi-supervised classification, and transfer learning.
Future works include extending \method for classifying other types of graphs.
% , including hypergraphs.
%\presecMargin

\vspace{-3mm}
\begin{credits}
\subsubsection{\ackname}
This work was supported by the National Research Foundation of Korea(NRF) funded by MSIT(2022R1A2C3007921).
This work was also supported by Institute of Information \& communications Technology Planning \& Evaluation(IITP) grant funded by the Korea government(MSIT) [No.2022-0-00641, XVoice: Multi-Modal Voice Meta Learning], [No.RS-2024-00509257, Global AI Frontier Lab], [No.RS-2021-II211343, Artificial Intelligence Graduate School Program (Seoul National University)], and [NO.RS-2021-II212068, Artificial Intelligence Innovation Hub].
The Institute of Engineering Research at Seoul National University provided research facilities for this work.
The ICT at Seoul National University provides research facilities for this study.
U Kang and Jinhong Jung are the corresponding authors.

% \subsubsection{\discintname}
% It is now necessary to declare any competing interests or to specifically
% state that the authors have no competing interests. Please place the
% statement with a bold run-in heading in small font size beneath the
% (optional) acknowledgments\footnote{If EquinOCS, our proceedings submission
% system, is used, then the disclaimer can be provided directly in the system.},
% for example: The authors have no competing interests to declare that are
% relevant to the content of this article. Or: Author A has received research
% grants from Company W. Author B has received a speaker honorarium from
% Company X and owns stock in Company Y. Author C is a member of committee Z.
 \end{credits}

% \vspace{-1mm}
\bibliographystyle{splncs04}
\bibliography{main}

\end{document}